# Development of Architectures for Internet Telerobotics Systems

**Riyanto Bambang**

School of Electrical Engineering and Informatics  
Bandung Institute of Technology  
e-mail: briyanto@lskk.ee.itb.ac.id

**Abstract**

This paper presents our experience in developing and implementing Internet telerobotics system. Internet telerobotics system refers to a robot system controlled and monitored remotely through the Internet. A robot manipulator with five degrees of freedom, called Mentor, is employed. Client-server architecture is chosen as a platform for our Internet telerobotics system. Three generations of telerobotics systems have evolved in this research. The first generation was based on CGI and two tiered architecture, where a client presents a Graphical User Interface to the user, and utilizes the user's data entry and actions to perform requests to robot server running on a different machine. The second generation was developed using Java. We also employ Java 3D for creating and manipulating 3D geometry of manipulator links and for constructing the structures used in rendering that geometry, resulting in 3D robot movement simulation presented to the users(clients) through their web browser. Recent development in our Internet telerobotics includes object recognition through image captured by a camera, which poses challenging problem, given the undeterministic latency of the Internet. The third generation is centered around the use of CORBA for development platform of distributed internet telerobotics system, aimed at distributing task of telerobotics system.

## 1 Introduction

Internet telerobotics is a new robotics field which attracted much attention among researchers in the last few years[3-5,8-15]. The term internet telerobotics refers to a robot system remotely controlled and monitored through the internet. The growing interest in this field is stimulated by the advancement of Internet, which provides access to various computing resources virtually from everywhere in the world. Increasing application of Internet as communication media for telerobotics system also comes from the fact that it uses standard communication protocol, and that the physical media is readily available for telerobotics application, eliminating the need for developing a dedicated, proprietary, and expensive communication system. Internet telerobotics differs from conventional telerobotics in that they are available, when desired, to the general public via the Internet. It allows people to actively participate in remote exploration in various applications. Areas where internet telerobotics is predicted to be useful include entertainment, telemanufacturing, telemedicine, teleoperation and mining. It has tremendous implications for education and training, allowing students and trainers to actively explore remote environments [5]. It also has implications for research as well, where laboratories can share access to expensive resources. There are problems, however, with the limited bandwidth of Internet for telerobotics system, particularly in regards to the visualization of robot movement from remote site, and with the security of Internet.

Early research on internet telerobotics was done by Goldberg and his research team. Their system consisted of a four-axis robot with a camera and air nozzle set up over a sandbox. Remote viewers could dig for objects buried in the sand from their Web browsers by positioning their mice and clicking on them. A team at the Univ. of Western Australia headed by Ken Taylor[5] developed internet telerobotics system consisted of six-axis robot manipulator and gripper which allows users to build structures with colored blocks. Telegarden, another installation developed by Goldberg's team, allows Web users to instruct a robot to plant and water seeds in a real garden.

This paper presents development and implementation of internet telerobotics systems. Client-server architecture supporting 2-tiered system is chosen as a platform for our Internet telerobotics system. Our development has resulted two generation internet telerobotics. First generation, developed in 1998, relies on CGI, while the second one, developed in 2000, is based on Java. User interface developed for the first generation internet telerobotics is based on HTML form and Java Applet. Two types of user interface in the client side are developed for the second generation. The first user interface is based on Java 3D, a relatively new development in internet telerobotics field. As far as the authors concern, utilization of this powerful tool has not been fully investigated in internet telerobotics. In our research, Java 3D is used for creating and manipulating 3D geometry of manipulator links, and for constructing the structures used in rendering that geometry. Alternative user interface was also developed using either remote robot image captured through a camera. Java 3D is particularly useful when implementing Internet telerobotics on networks with relatively slow bandwidth. The third generation is centered around the use of CORBA for development platform of distributed internet telerobotics system, aimed at distributing task of telerobotics system.

## 2. Architecture of First Generation Telerobotics

Architecture of the first generation of our telerobotics system is centered eround 2-tiered architecture concepts where a client presents a GUI (*Graphical User Interface*) interface to the user. The system utilizes the operator's data entry and actions to perform requests to robot server running on a different machine. However, it does not fully support the concepts of 2-tiered system since a large portion of resources used resides on the server. In addition, the robot simulation is not presented to the clients.

The first generation of internet telerobotics has the following features and capabilities :
- The system is capable to control robot manipulator with five degrees of freedom through web based application.
- The remote user provides set point for robot manipulator position
- Subsystem used by remote operator does not depend on the operating system of the client (*platform independence*).
- The system obtains feedback position data within a specified time duration and send it to user monitor.

The telerobotics system uses HTTP services through HTTP server application running under the chosen operating system. The networks used should allow direct connection on port 80 for HTTP service without passing proxy. The connection should have minimum bandwidth of 9600 Bps to allow speedy response. The feedback data is received after robot reaches their desired position, after which its visualization is performed. The resulting architecture is depicted in Figure 2.1. The telerobotics system comprises of 3 main subsystems : robot manipulator and interface SB board, robot server PC and remote (client) PC attached to the internet. Robot server PC consists of robot actuating and feedback data sampling sub-system, Common Gateway Interface (CGI), and HTTP server.

Data and information flows in a user request cycle from an instance when the user provides desired position through instance when he/she receives the actual position of the robot, is directed as follows (see Figure 2.2):
1. User provides desired position of each axes of the robot through web browser. Web browser sends the data to web server.

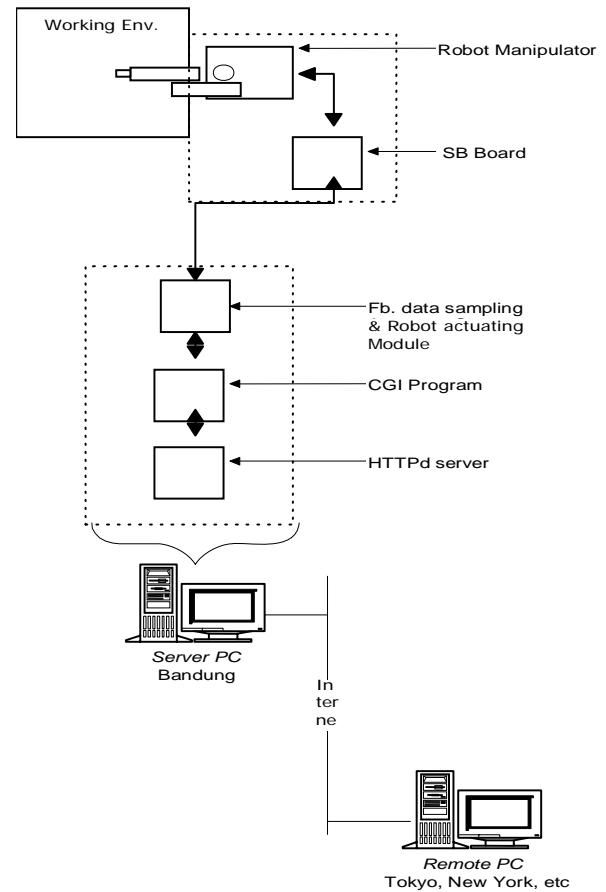

Figure 2.1. General architecture of 1st generation of internet telerobotics

2. Web server initiates CGI program and provides the data.
3. CGI Program provides the input data to robot actuation and robot data sampling modules.
4. Robot actuation module sends the set point to SB Board through serial RJ45 interface.
5. The SB Board sends the actual robot position to robot actuating & data sampling module.
6. The data is sent through CGI.
7. CGI Program sends the data received to web server.
8. Web server sends the actual position data to web browser to be displayed on the user monitor

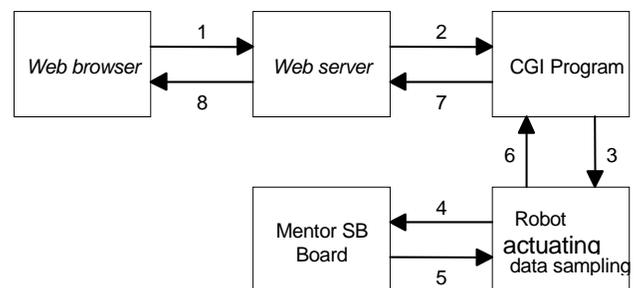

Figure 2.2. Data flow in Internet telerobotics

One of the important issues in Internet telerobotics is time delay due to undeterministic nature of TCP/IP

protocol. In our Internet telerobotics system, an operator should wait for a response to a previous command before he/she can issue the next command. The corresponding time delay (telerobotics response) is defined by

$$t_r = t_p + \frac{\sum D}{v_l} + t_C$$

where :

$t_r$ = telerobotics time response

$t_p$ = time for request processing

$\sum D$ = the amount of data sent and received

$v_l$ = connection link speed

$t_C$ = time required to initiate communication

Among the parameters involved in the above equation, we can adjust only time required to process request and the amount of data sent and received, whereas other parameters tend to be fixed. Thus if it is required to decrease telerobotics time response, the time-duration required to process a request and/or the data sent and received should be reduced.

The main function of CGI program is to receive data from web server consisting of set point data for each robot axes, and in turn send it to robot actuating module. It also commands position feedback data sampling module to read data to be sent to web server. CGI program is run by web server once a request is received to execute it. When the request is completed, CGI program terminates and freezes content of memory it used. Data sent by web server to CGI or vice-versa has string format, whereas processing by the robot actuating module is performed using integer type. Therefore, CGI program also functions as data type conversion for entering and leaving information flow.

Data communication between server and client in telerobotics system relies on HTTP protocol. Data communication based on this protocol is performed between web browser in the client side and web server in the server side. Port 80 is used by HTTP protocol. However, there is an exception when using user interface based on Java, since in this case web browser does not control data transmission to the server, but instead used as Java interpreter. On the other hand Java applet serves mainly as remote application which performs data communication to web server so as to gain access to CGI program. The client and server software component for telerobotics using form is shown in Figure 2.3, while that with java based interface is depicted in Figure 2.4.

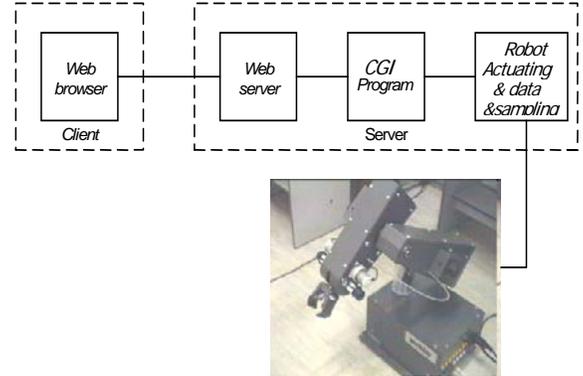

Figure 2.3 Block diagram of internet telerobotics with form based user interface

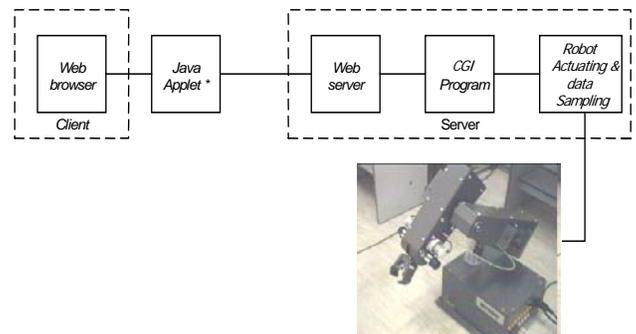

Figure 2.4 Block diagram of internet telerobotics with Java applet based user interface

By employing Java technology, data communication between Java applet application and web server can be carried out using URL(*Uniform Resource Locator*) class included in java.net package. URL stores two important information : protocol marking and resource. URL class allows us to read and write data at a resource address on Internet. Before Java Applet transmits data, it opens connection to a URL containing designated CGI program. Once connected, data from URL is read through input stream. Information transmission from web browser on the client side to web server on the server side is performed using Post method included in form. The steps required to transmit data is basically similar to the data transmision from Java Applet to CGI porgram.

## 3. Architecture of Second Generation of Telerobotics

In the second generation of internet telerobotics, a client-server system with shared control mode on each server and client is constructed. This system is divided into two sub-system : Client and Server. The term client-server architecture is a general description of a networked system where a client program initiates contact with a separate server program (possibly on a different machine) for a specific function or purpose. The client is located in the position of the requester for the service provided by the server. This system is improvement over our first generation internet telerobotics which used standard web application using Common Gateway Interface (CGI) as a core of the system. In the first generation system, visualization window was not developed due to limitation in CGI programming.

In the second generation, application logic is typically tied to the client application and a 'heavy' network process is required to mediate the client-server interaction. Such a system fully supports the concepts of two tiered system, where Robot Server and Web Server reside on the same machine.

In a fully 2-tiered system, a distributed control system is constructed to reduce response time for completing a task. By using a separate system, a client and a server work concurrently. In comparison to the first generation system, such a mechanism will reduce server's load while other site will increase client's load. However, since in 2-tiers architecture client and server are implemented on different machines, the loads of client and server are balanced.

The proposed mode of telerobotic system has been implemented using Java and leads to a distributed telerobotic system with the general architecture shown in Figure 3.1. In this architecture, when multiple users are trying to operate the robot from possibly different places, while another user is operating the robot, their requests are placed in a qeuee. A first-in first-serve mechanism is introduced for multiple user requests. The request is processed by Java Server by sending the command to the robot while receiving feedback position. In this research, we used Java user interface based on 3-D simulation.

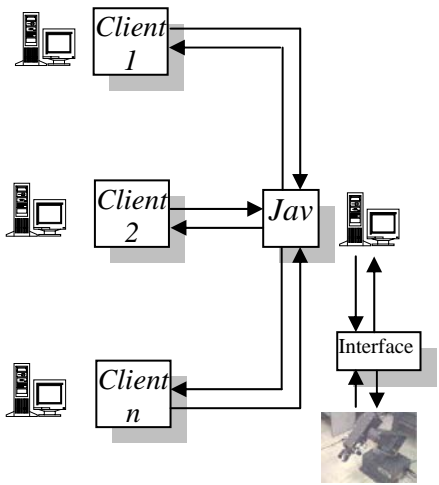

Figure 3.1. Second generation internet telerobotics

To implement 3D simulation, geometry of all robot axes and position are measured. To produce a fairly smooth simulation, model of each axes and their accecories have to be created, along with their behaviour. In this simulation, all data is accepted from server as a serial data of current robot position.

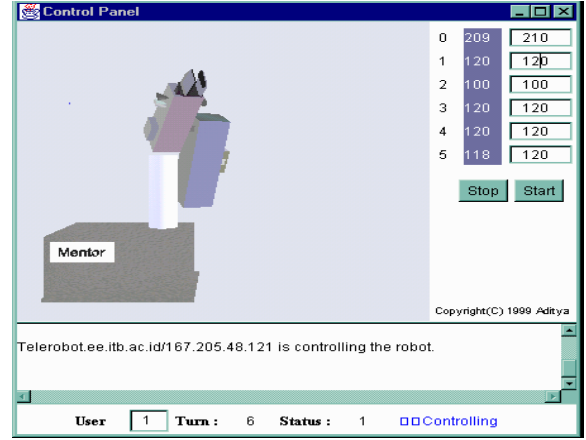

Figure 3.2 Feature of Java 3D simulation

To distinguish type of command and feedback data from those of others such as data of client's number which currently access to the server, IP or host name that currently goes in or out, user status etc., a symbol is used in front of each serial data. The feature of the simulation is shown in Figure 3.2. The components in the client site for receiving data include values conversion, status checking, data splitter and buffer, while for transmitting data consist of status checking and serialization.

## 4. Implementation Results

We performed two types of testing. The first is shape and motion comparison between simulation features and real motion. The results (not presented for lack of space) shows that the 3D simulation approximates well the robot real motion. We found a problem, however, with gripper motion especially when it is in a difficult position and performs a complicated motion.

The second test is time response. We choose two setpoint data to check time response needed to complete an assignment. We also tested network time response of packet ICMP (Internet Control Message Protocol) using *ping* software in Unix operating system. We performed time response testing in local computer (client and server reside on the same machine), in local network (client and server reside on different machines, but remain in same LAN), and in Internet (client and server are in the different LANs). The average results are shown in Table 1. From these results, we found that the ICMP time response increases as the operator moves from Local to LAN to Inter-LAN. However, time responses for 1 cycle task are almost the same for local, LAN and Inter-LAN connections. We found that the time response for local connection is smaller than that of LAN connection. It is caused by client load that should be engaged in the server machine.

**Table 1. Experimental time response results**

| Test Condition | ICMP response (ms) | Time response for 1 cycle task (s) |
|---|---|---|
| Local | 1 | 3 |
| LAN | 1.2 | 3.1 |
| Inter-LAN | 7.75 | 3.1 |

3D simulation of robot arm is shown in Figure 4.1, while its real action is shown in Figure 4.2. The simulation shown in Figure 4.1 is presented to the clients, while the real action of robot is located at the server (refer to Fig. 3.1).

Finally, it should be pointed out that server load is only one of the factors affecting telerobotic time response performance. Other factors include network traffic (ICMP response), software delay, and hardware delay. However, the Java 3D API is found to be useful in providing a speedy time response.

### 5. Third Generation Architecture

The third generation architecture system is recently developed, aimed at distributing tasks for telerobotics system. The architecture is based on CORBA(Common Object Request Broker Architecture), a middleware architecture specification commonly employed for distributed system. For a real-time application, such as telerobotics system, however, common implementation of CORBA may not be applied. Therefore, we choose TAO, a real-time ORB end system implementation. TAO consists of communication interface, operating system's I/O subsystem, communication protocol, and CORBA middleware components (see Figure 5.1). TAO also implements Audio/Video streaming in CORBA, which is particularly useful for remote monitoring of a robot through camera.

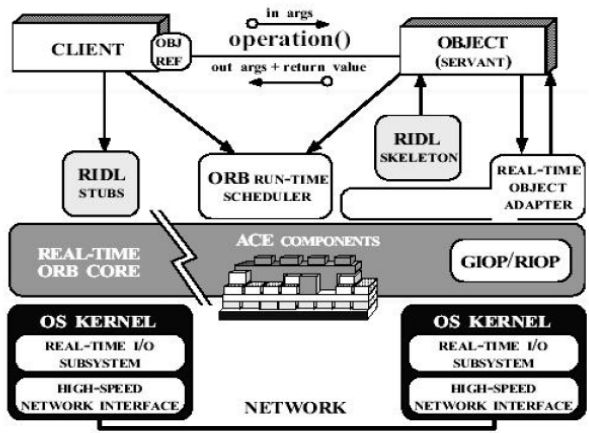

Figure 5.1 TAO Implementation of Real-Time CORBA

Architecture of distributed telerobotics using CORBA is shown in Figure 5.1. Such an architecture allows operators to jointly and in a distributed manner provide commands to control the remote robot.

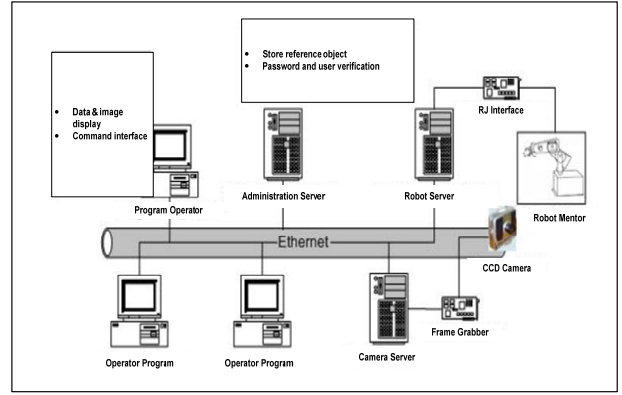

Figure 5.1 Architecture of Distributed Telerobotics Using CORBA

### 6. Conclusion

This paper presents our research on internet telerobotics systems. The system has evolved from internet telerobotics based on CGI to one relying on Java, and CORBA. Client-server architecture has been developed as a platform for telerobotics system. Recent research includes utilization of camera to capture image around working objects, pattern recognition based on neural networks which is integrated in telerobotics system. To address important issue of communication delay inherent in TCP/IP protocol, particularly for transmitting image captured by a camera from the server to the client, we recently investigate the utilization of Real-time Transport Protocol (RTP). Other important issue in Internet telerobotics that has not been addressed in this paper is security. This is left for future research.

Java 3D API is an efficient tool for creating and manipulating 3D geometry of manipulator links in an internet telerobotics system and for constructing the structures used in rendering that geometry. It provides a way to improve performance of the system from the view point of visualization and time response, and is therefore effective in a low speed network. The internet based telerobotic architecture developed in this research along with its Java 3D simulation is expected to be useful in the future applications of teleoperation.